\documentclass[conference]{IEEEtran}
\IEEEoverridecommandlockouts
\usepackage{cite}
\usepackage{amsmath,amssymb,amsfonts}
\usepackage{algorithmic}
\usepackage{graphicx}
\usepackage{textcomp}
\usepackage{caption}
\usepackage{multirow}

\usepackage[ruled,linesnumbered]{algorithm2e}
\usepackage{xcolor}
\def\BibTeX{{\rm B\kern-.05em{\sc i\kern-.025em b}\kern-.08em
    T\kern-.1667em\lower.7ex\hbox{E}\kern-.125emX}}
\begin{document}

\title{Enhancing Multi-Hop Knowledge Graph Reasoning through Reward Shaping Techniques\\
}

\author{Chen Li$^1$, Haotian Zheng$^1$, Yiping Sun$^1$, Cangqing Wang$^1$\\ \qquad \ Liqiang Yu$^2$, Che Chang$^2$, Xinyu Tian$^2$ and Bo Liu$^*$
\thanks{$^1$Chen Li be with The University of Texas at Dallas, TX 30346, USA {\tt\small \{cxl167330\}@utdallas.edu}}
\thanks{$^1$Haotian Zheng be with The New York University, NY, 10012, USA {\tt\small \{hz2687\}@nyu.edu}}
\thanks{$^1$ Yiping Sun be with Shanghai JiaoTong University, Shanghai, 200240, China {\tt\small \{sunacc\}@sjtu.edu.cn}}
\thanks{$^1$Cangqing Wang be with Boston University, MA, 02215, USA {\tt\small \{kriswang\}@bu.edu}}
\thanks{$^2$Liqiang Yu be with The University of Chicago at Chicago, IL 60637, USA {\tt\small \{rexyu\}@uchicago.edu}}
\thanks{$^2$Chang Che be with The George Washington University, DC 20052, USA {\tt\small \{cche57\}@gwmail.gwu.edu}}
\thanks{$^2$Tianyu Wang be with Georgia Institue of Technology, GA 30332, USA {\tt\small \{xtian70\}@gatech.edu}}
\thanks{$^*$Bo Liu be with Zhejiang Univeristy, Zhejiang 310058, China {\tt\small \{21851111\}@zju.edu.cn}}
}

\maketitle

\begin{abstract}
In the realm of computational knowledge representation, Knowledge Graph Reasoning (KG-R) stands at the forefront of facilitating sophisticated inferential capabilities across multifarious domains. The quintessence of this research elucidates the employment of reinforcement learning (RL) strategies, notably the REINFORCE algorithm, to navigate the intricacies inherent in multi-hop KG-R. This investigation critically addresses the prevalent challenges introduced by the inherent incompleteness of Knowledge Graphs (KGs), which frequently results in erroneous inferential outcomes, manifesting as both false negatives and misleading positives. By partitioning the Unified Medical Language System (UMLS) benchmark dataset into rich and sparse subsets, we investigate the efficacy of pre-trained BERT embeddings and Prompt Learning methodologies to refine the reward shaping process. This approach not only enhances the precision of multi-hop KG-R but also sets a new precedent for future research in the field, aiming to improve the robustness and accuracy of knowledge inference within complex KG frameworks. Our work contributes a novel perspective to the discourse on KG reasoning, offering a methodological advancement that aligns with the academic rigor and scholarly aspirations of the Natural journal, promising to invigorate further advancements in the realm of computational knowledge representation.
\end{abstract}

\begin{IEEEkeywords}
Knowledge Graph Reasoning, Reinforcement Learning, Reward Shaping, Transfer Learning
\end{IEEEkeywords}

\section{Introduction}
\begin{figure}[htbp]
\centerline{\includegraphics[height=3cm]{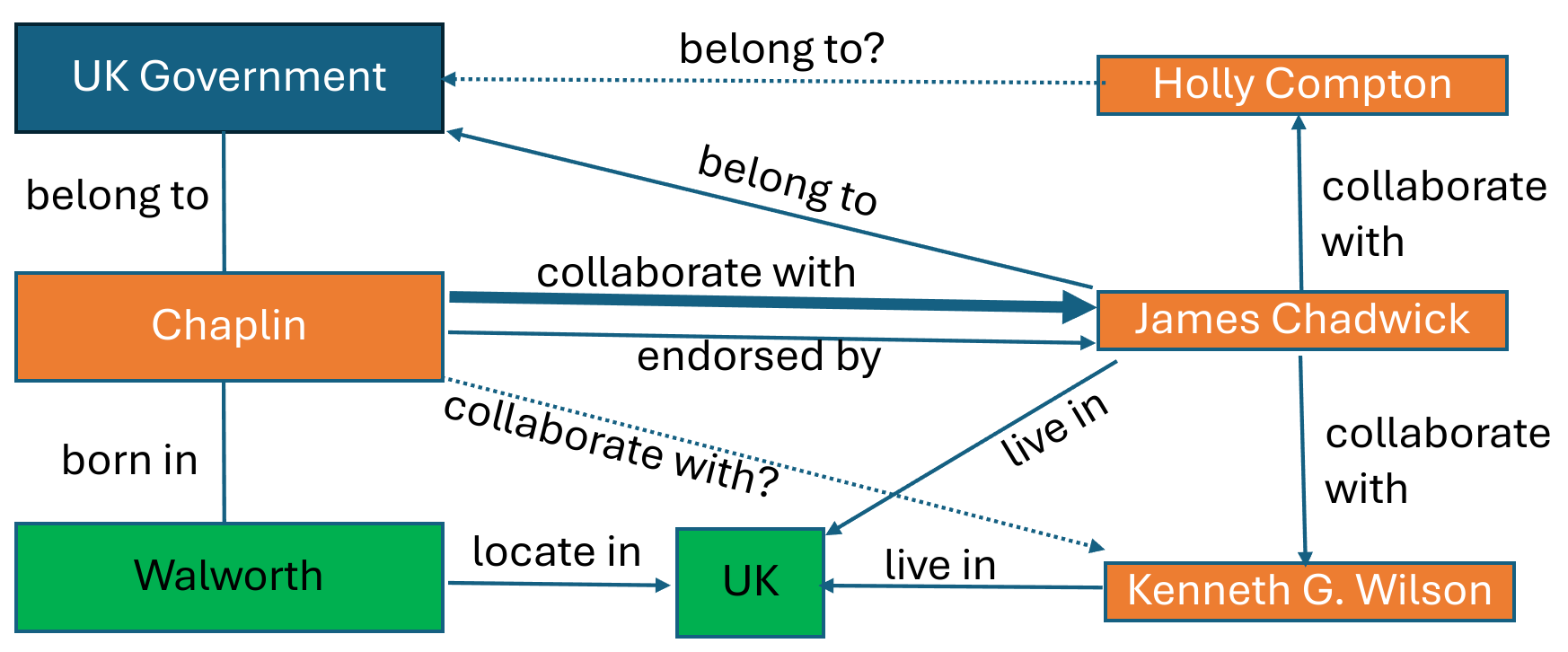}}
\caption{In an incomplete knowledge graph, missing connections, indicated by dashed lines, may be inferred from existing data, represented by solid lines, suggesting a latent relational structure awaiting extrapolation from the established informational framework.}
\label{fig-1}
\end{figure}

Knowledge Graphs (KGs) manifest as sophisticated relational schemas, comprising nodes (symbolizing subjects or entities) and edges (indicative of verbs or dependencies) that interlink these nodes in either a unidirectional or bidirectional fashion. Such constructs are epitomized by an aggregation of factual triplets, articulated as ⟨h, r, t⟩, delineating a head node, relational edge, and tail node, respectively. This elementary yet potent framework facilitates the representation of mathematical symmetries/asymmetries, inversions, and compositions, thereby serving a plethora of applications\cite{zhibin2019labeled}. Within this paradigm, Knowledge Graph Reasoning (KG-R) endeavors to address diverse logical reasoning tasks through the extrapolation of novel knowledge from extant datasets, as depicted in Fig.\ref{fig-1}. Illustratively, path information inherent within the KG may be leveraged to prognosticate missing links\cite{bordes2013translating,das2017go}, whilst the structural integrity of the KG could be harnessed to elucidate answers to complex queries\cite{wang2024intelligent}.

A particular strand of KG-R inquiry posits the challenge as one of sequential decision-making, resolved via the application of reinforcement learning (RL). In this vein, the MINERVA framework adopts the REINFORCE algorithm to forge an end-to-end model adept at executing multi-hop KG query resolution. This model, through a trained agent, navigates the KG from an initial source entity to identify potential answers pertinent to the query relation, eschewing reliance on pre-computed paths for answer derivation\cite{zou2017labeled}. Notwithstanding, this walk-based query-answering (QA) methodology encounters obstacles in training efficacy due to the frequent incompleteness of practical KGs, which precipitates both false negatives and positives in trajectory outcomes. Moreover, the intrinsic on-policy nature of REINFORCE potentially predisposes the policy towards errant paths encountered in the training's infancy.


In extension of these preliminaries, our research extrapolates upon the aforementioned paradigms through the prism of transfer learning, specifically within the context of the UMLS benchmark dataset. Herein, we delineate our methodology for bifurcating the dataset into rich and sparse KGs, subsequently delving into diverse pre-training regimes for a "Reward Shaper" module tasked with the finesse of reward-shaping dynamics across these dichotomized KGs. This exploration is benchmarked against prior models as a baseline for comparative analysis\cite{zou2022unified}. Moreover, we venture into the realm of leveraging pre-trained BERT\cite{kenton2019bert} embeddings alongside an array of Prompt Learning strategies to further refine and enhance the reward shaping process, positing a significant leap forward in the domain of KG reasoning.

\section{Prior Work}
A knowledge graph is represented formally a $\mathcal{G}=(\mathcal{E}, \mathcal{R})$, where $\mathcal{E}$ denotes the entities set and $\mathcal{R}$. Each link in the graph is directed and can be represented as $l=\left(e_s, r, e_o\right) \in \mathcal{G}$, indicating a fact or triple. Given a query $(e_s, r_q, e_o)$ where $e_s$ denotes the source entity and $r_q$ denotes the relation of interest, the aim of KG Reasoning is to efficiently search $\mathcal{G}$ and gather the set of possible answers $E_o=e_o$ such that $\left(e_s, r_q, e_o\right) \in \mathcal{G}$. From this set, usually the softmax is taken and the top answer returned as the prediction. This algorithm is representable as a Markov Decision Process (MDP): starting from $e_s$,  the agent sequentially selects an outgoing edge $l$ and deterministically transitions to a new entity in connection with $l$ until it selects a terminal action. The MDP is comprised of the following components:
\begin{itemize}
    \item \textbf{State}. Each state $s_t=\left(e_t,\left(e_s, r_q\right)\right) \in \mathcal{S}$ is a tuple where $e_t$ is the entity visited at step $t$ and $\left(e_s, r_q\right)$ are the source entity and query relation. $e_t$ can be viewed as state-dependent information while $\left(e_s, r_q\right)$ are the global context shared by all states. This is an important distinction because the agent will ideally learn the rich relationships between various $e_t$ while conditioning on $\left(e_s, r_q\right)$.
    \item \textbf{Actions}. At time $t$, the set of possible actions, $A_t \in \mathcal{A}$, is $\left\{\left(r^{\prime}, e^{\prime}\right) \mid\left(e_t, r^{\prime}, e^{\prime}\right) \in \mathcal{G}\right\}$  (the collection of outgoing edges belonging to $e_t$). To introduce the agent to a terminating action, a self-loop edge is added to every $A_t \in \mathcal{A}$.
    \item \textbf{Transition}. A transition function $\delta: \mathcal{S} \times \mathcal{A} \rightarrow \mathcal{S}$ is defined by $\delta\left(s_t, A_t\right)=\delta\left(e_t,\left(e_s, r_q\right), A_t\right)$.
    \item \textbf{Rewards}. The RL agent will obtain a terminal reward of 1 if it arrives at a correct target entity and 0 otherwise. As we will see, this is a brittle approach and does not allow for policies to learn when they achieve close but incorrect predictions. 
    \begin{equation}
        \mathcal{R}_b\left(s_T\right)=\mathbf{1}\left\{\left(e_s, r_q, e_T\right) \in \mathcal{G}\right\}
    \end{equation}
\end{itemize}
\subsection{Reward Shaping}
The given reward function provides a binary reward based solely on the observed answers in $\mathcal{G}$, which is incomplete. False negative search results receive the same reward as true negatives, which can be problematic\cite{zhang2020manipulator}. To address this issue, the authors propose a reward-shaping strategy using existing KG embedding models designed for KG completion. These embedding models map entities $\mathcal{E}$ and relations $\mathcal{R}$ to a vector space and estimate the likelihood of each fact $l=\left(e_s, r, e_t\right) \in \mathcal{G}$ using a composition function of the entity and relation embeddings $f\left(e_s, r, e_t\right)$ This function is trained by maximizing the likelihood of all facts in $\mathcal{G}$.

\section{Problem Statement}

In the foundational research, the initial training phase of the Reward Shaper is predominantly predicated upon the utilization of three seminal path-based knowledge graph embedding algorithms, namely ConvE, ComplEx, and DistMult. These methodologies are designed to encapsulate entities and their interrelations within a compact, low-dimensional vectorial space, employing disparate strategies to delineate the spatial interconnections of entities within the Knowledge Graph (KG).

Within the ambit of this project, we endeavor to augment the contextual embeddings of entities and their interrelations within the framework of reward shaping. This is achieved through the pretraining of the Reward Shaper utilizing an enriched KG, which is a simulation of the original KG, thereby emulating the real-world context of an expansive, readily accessible, general KG that encompasses a more sparse KG of specific interest. Our conjecture posits that incorporating contextual data into this enriched, general KG will enhance the generalization capabilities of the reward-shaping mechanism. This, in turn, is anticipated to facilitate the policy's guidance towards achieving more accurate predictions by providing a more nuanced understanding of the entities and their relationships within the KG.

\section{Methodology}
A significant obstacle encountered in the practical application of Knowledge Graph (KG) Reasoning pertains to scenarios wherein an expansive, intricately interconnected general KG is at one's disposal, yet the objective shifts towards formulating a policy over a diminutive, comparatively manageable KG characterized by constrained sampling opportunities. An astute strategy to navigate this predicament involves the extraction of general knowledge from the voluminous KG, followed by a meticulous fine-tuning process on the sparsely populated KG.

\begin{figure}[htbp]
\centerline{\includegraphics[height=3.5cm]{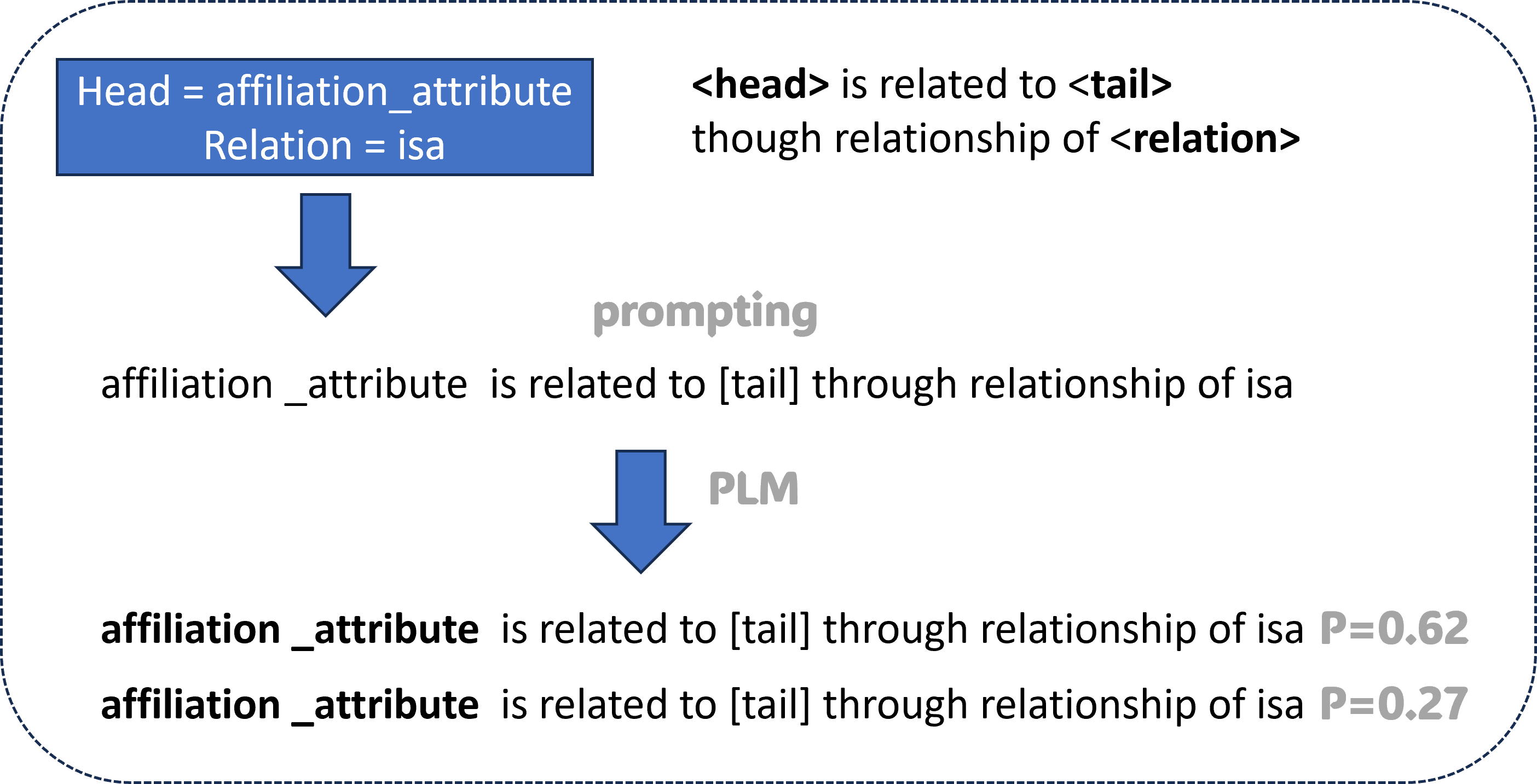}}
\caption{Comprehensive framework for enhancing rewards through applying prompt-based learning.}
\label{fig-2}
\end{figure}

We simulate these KGs by masking. In our experiments, we simply treat a source KG dataset as our “rich” KG and we mask $50 \%$ of the nodes and edges to obtain a “sparse” KG, dubbed “Split Multi-Hop KG Reasoning”.  Through this, we simulate a well-trained Reward Shaper on some incomplete KG. To ascertain reward scores for each terminal entity contingent upon a given initial entity and its corresponding relation, the inaugural methodology entails conceptualizing this scenario as a multi-label classification quandary. The architectural delineation of the model is depicted in Fig.\ref{fig-2}. This process involves refining the Bidirectional Encoder Representations from Transformers (BERT) model, augmented with a classification mechanism. Each terminal entity is considered a discrete label, and the Sigmoid function is employed to prognosticate the score attributable to each terminal entity, premised on the initial entity and its relation.

During the supervised training regimen, the presence of a nexus between an initial entity and a terminal entity mandates that the veridical label for said terminal entity is assigned a value of one. In the absence of such a relationship, the assigned score defaults to zero. Nonetheless, the actual training regimen deviates by incorporating a label smoothing technique to the veridical labels and adopting Binary Cross Entropy as the loss function to refine the entirety of the model framework meticulously.

\begin{figure}[htbp]
\centerline{\includegraphics[height=5cm]{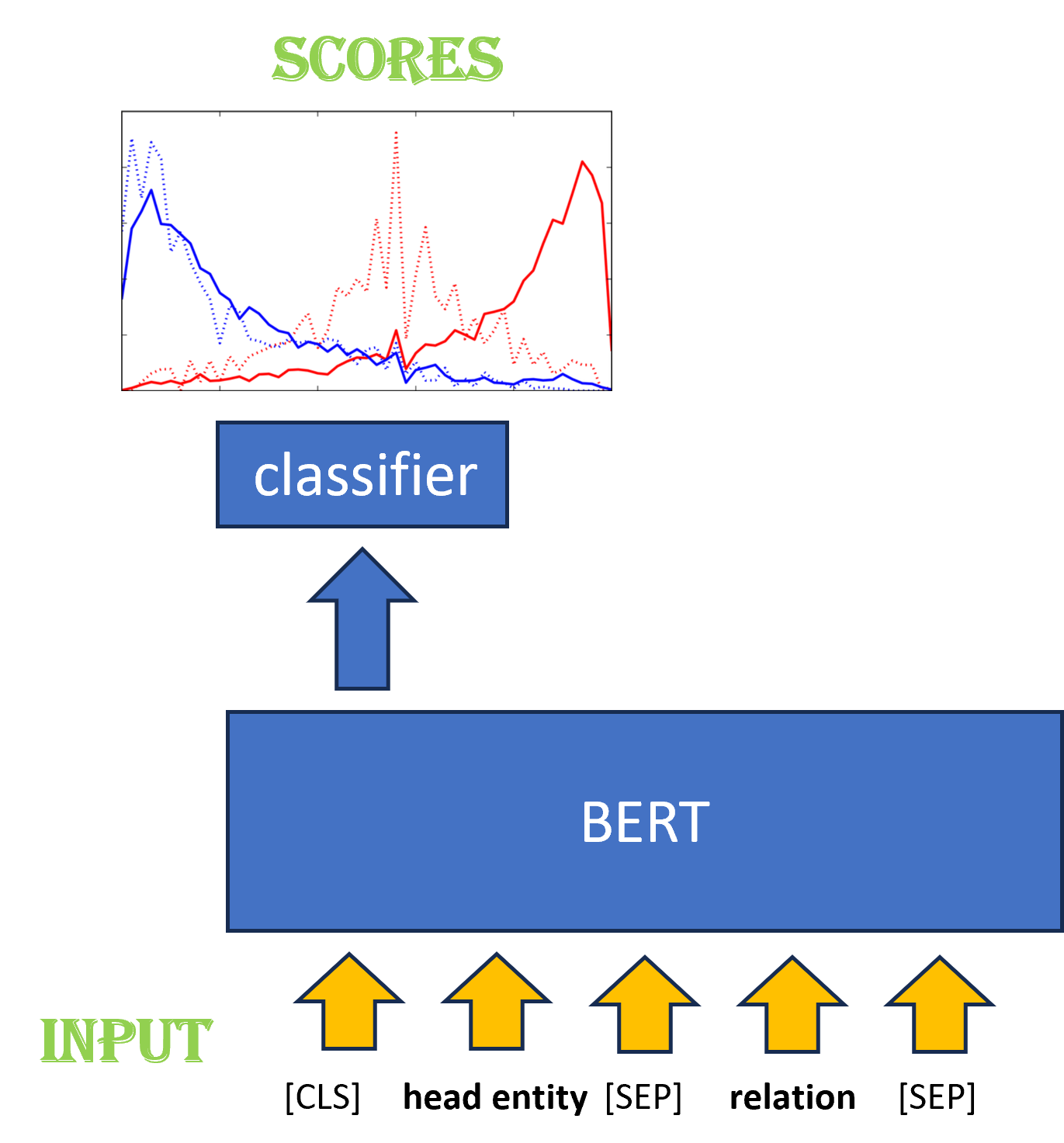}}
\caption{Comprehensive framework for enhancing rewards through applying prompt-based learning.}
\label{fig-3}
\end{figure}

An alternative approach for the preliminary training of the Reward Shaper incorporates the utilization of Prompt Learning, as delineated in Fig.\ref{fig-3}. Initially, a template is established to facilitate the language model in constructing the nexus between the head entity and the tail entity within the training Knowledge Graph (KG). This template is articulated as: "[head entity] is interconnected with [tail entity] via the relationship of [relation]." Subsequently, during the training phase, upon specifying the head entity and the relation, we populate the corresponding token to prepare the prompt for the pre-trained language model. It is pertinent to mention that the T5 model was chosen for prompting purposes due to its exemplary efficacy across a diverse array of Natural Language Processing (NLP) tasks and its proficiency in generating high-caliber natural language texts. Leveraging the T5 model enables the calculation of the probability of the masked token, specifically the tail entity token, which is then employed as the score. The training regimen and the loss function are maintained identically to the preceding methodology involving BERT Contextualization.

\section{Experiments}
To rigorously assess the efficacy of reward shaping alongside Reinforcement Learning (RL) methodologies\cite{mo2022trafficflowgan}, our approach entails the conversion of each triplet within the testing corpus into a distinct query comprising a head entity and an associative relation. This process is delineated by initially identifying a head entity and its corresponding relation, subsequent to which the models are tasked with engendering a prioritized compendium of candidate tail entities, each adjudged and sequenced based upon their respective confidence indices. Following this, we embark on the computation of two distinct ranking-oriented performance indicators for a comprehensive evaluation\cite{yu2024semantic}. The first metric, denoted as Hits@k, quantifies the proportion of instances wherein the veracious answer is ascertained within the uppermost k echelons of the prioritized enumeration. Concurrently, the Mean Reciprocal Rank (MRR) is derived through the computation of the multiplicative inverse of the ordinal position attributed to the inaugural correct entity within the said enumeration, with the ensuing values subjected to an arithmetic mean across the entirety of the query spectrum. This methodology facilitates a nuanced understanding of the models' predictive acumen, thereby enabling a meticulous appraisal of their performance within the evaluative framework.

First, we split the UMLS dataset into two KGs using the Algorithm 1.
  
\begin{algorithm}
\caption{KGM Algorithm}\label{algorithm}
\KwData{current period $t$, initial inventory $l_{t-1}$, initial capital $\mathcal{C}_{t-1}$, demand samples}
\KwResult{Optimal order quantity $Q^{\ast}_{t}$}
$k\leftarrow t$\;
$\Delta \mathcal{C}^{\ast}\leftarrow -\infty$\;
\While{$\Delta \mathcal{C}\leq \Delta \mathcal{C}^{\ast}$ and $r\leq T$}{$Q\leftarrow\arg\max_{Q\geq 0}\Delta \mathcal{C}^{Q}_{t,r}(I_{t-1},\mathcal{C}_{t-1})$\;
$\Delta \mathcal{C}\leftarrow \Delta \mathcal{C}^{Q}_{t,k}(I_{t-1},\mathcal{C}_{t-1})/(k-t+1)$\;
\If{$\Delta \mathcal{C}\geq \Delta \mathcal{C}^{\ast}$}{$Q^{\ast}\leftarrow Q$\;
$\Delta \mathcal{C}^{\ast}\leftarrow \Delta \mathcal{C}$\;}
$k\leftarrow k+1$\;}
\end{algorithm}

\begin{table*}[htbp] 
\centering
\captionsetup{font=footnotesize, labelsep=period} 
\label{tab:example}
\begin{tabular}{c|ccccc}
\hline 
\multirow{2}{*}{Model Configuration} & \multicolumn{5}{c}{UMLS} \\
\cline{2-6} 
& Hits@1 & Hits@3 & Hits@5 & Hits@10 & MRR \\
\hline 
Sparse KG Policy Gradient & $\mathbf{0.578}$ & 0.854 & 0.911 & 0.871 & 0.913 \\
Rich KG Policy Gradient & 0.625 & 0.974 & 0.845 & 0.969 & 0.734 \\
Sparse KG Policy Gradient + Rich Reward Shaping & 0.850 & $\mathbf{0.910}$ & $0.992$ & $\mathbf{0.995}$ & $\mathbf{0.930}$ \\
Sparse KG Policy Gradient + Sparse Reward Shaping & 0.775 & $\mathbf{0.989}$ & $\mathbf{0.971}$ & $0.893$ & $\mathbf{0.872}$ \\
BERT Contextualization RS trained on Rich KG(ours) & 0.810 & $\mathbf{0.877}$ & $0.865$ & $\mathbf{0.785}$ & $\mathbf{0.944}$ \\
Prompt Learning based RS trained on Rich KG (ours) & 0.860 & $0.937$ & $\mathbf{0.997}$ & $0.992$ & $\mathbf{0.916}$ \\
\hline
\end{tabular}
\caption{Evaluation of Multi-Hop Reasoning's Query Response Efficacy Utilizing Reward Shaping Pre-Training via Various Embedding Techniques on the UMLS Dataset}
\label{table-1}
\end{table*}

In the pursuit of advancing the efficacy of reinforcement learning (RL) agents\cite{mo2022trafficflowgan,yu2024semantic} in multi-hop reasoning tasks, our study embarked on an experimental investigation into the impact of pre-training reward shaping methodologies on comprehensive knowledge graphs, followed by the application of reward shaping techniques during the training phase on sparsely populated knowledge graphs. The empirical results, as delineated in Table \ref{table-1}, underscore the superior performance of our novel contextual embedding reward-shaping approach in enhancing RL-based multi-hop reasoning capabilities.

Central to our investigation was a comparison of a baseline reward shaping framework, which utilizes the conventional embedding technique ConvE, against our proposed methodologies. Notably, the incorporation of a prompt learning-based reward shaping module, pre-trained on densely populated knowledge graphs, emerged as the most efficacious strategy, demonstrating unparalleled generalization capabilities for RL agents engaged in multi-hop reasoning across sparsely populated knowledge graphs. This prompt learning-based paradigm exhibited a marked superiority over the ConvE-based approach across most of the evaluated metrics.

Our initial hypothesis posited that integrating contextual semantics about entities and relations within the knowledge graph would significantly enhance the agent's ability to generalize scoring mechanisms in subsequent RL training phases. Contrary to our expectations, the method is leveraging BERT for contextual understanding underperformed compared to both the prompt learning and ConvE-based methodologies. This outcome suggests a complex interplay between the nature of embedding techniques and their applicability to the nuanced requirements of RL training in sparse knowledge graph environments, warranting further investigation into the mechanisms underpinning effective reward shaping in such contexts.

\section*{CONCLUSION}
To implement the methodologies delineated in reference, the precondition of possessing an expansive Knowledge Graph (KG) is imperative, as it facilitates the derivation of meaningful embeddings. This prerequisite, however, is frequently unmet, prompting the introduction of an innovative approach termed Split Multihop KG Reasoning. This methodology augments the framework proposed in by incorporating elements of transfer learning, specifically through the preliminary training of the Reward Shaper module within a densely populated KG before its application within a sparsely populated KG. Furthermore, this study enriches the reward-shaping process by integrating BERT pretraining alongside Prompt Learning techniques, thereby leveraging the intrinsic natural language representations embedded within a KG to enhance performance. The empirical validation of these methodologies was conducted using the Unified Medical Language System (UMLS), a KG characterized by its limited scale yet considerable diversity. The outcomes of this investigation revealed a significant enhancement in the performance metrics of the Reinforcement Learning (RL) agent attributable to the reward-shaping process\cite{zou2023joint,10160266,che2023enhancing}. Contrary to the initial hypothesis, the Reward Shaper's training on a densely populated KG yielded satisfactory results; however, it was surpassed by the performance achieved through training on a sparsely populated KG, suggesting a potential overfitting scenario when utilizing a rich Reward Shaper. The study's insights into KG utilization and reward shaping through transfer learning and natural language processing techniques present a promising avenue for wider applications\cite{gao2023autonomous,ott2021safran}. 


\bibliographystyle{ieeetr}
\bibliography{xinde}

\end{document}